\documentclass[conference]{IEEEtran}
\IEEEoverridecommandlockouts
\usepackage{cite}
\usepackage{amsmath,amssymb,amsfonts}
\usepackage{algorithmic}
\usepackage{graphicx}
\usepackage{textcomp}
\usepackage{xcolor}
\usepackage{float}
\usepackage{booktabs}
\usepackage{subfigure}

\def\BibTeX{{\rm B\kern-.05em{\sc i\kern-.025em b}\kern-.08em
    T\kern-.1667em\lower.7ex\hbox{E}\kern-.125emX}}

\newcommand{\fid}{FID$\downarrow$}
\newcommand{\kid}{KID$\downarrow$}

\newcommand{\loss}{L}

\begin{document}

\title{Looks Like $\mathcal{M}$agic: Transfer Learning in GANs to Generate New Card Illustrations \\
}

\author{\IEEEauthorblockN{Matheus K. Venturelli, Pedro H. Gomes, Jônatas Wehrmann}
\IEEEauthorblockA{\textit{Pontifical Catholic University of Rio de Janeiro (PUC-Rio)} \\
Rio de Janeiro, Brazil \\
\{mventurelli,pbarroso\}@inf.puc-rio.br, wehrmann@puc-rio.br}
}

\def\fancymagic{$\mathcal{M}$\textsc{agic}}

\def\magic{\fancymagic\textsc{: The Gathering}}
\def\gan{\textsc{MagicStyleGAN}}
\def\ganada{\textsc{MagicStyleGAN-ADA}}
\def\dataset{\textsc{MTG}}
\def\generator{\mathcal{G}}
\def\discriminator{\mathcal{D}}
\def\mapping{\mathcal{F}}
\def\noisespace{\mathcal Z}
\def\noise{\mathbf{z}}
\def\projnoise{\mathbf{w}}
\def\projnoisespace{\mathcal W}
\def\normal{\mathcal N(\mu, \sigma^{2})}
\def\imagedist{\mathcal I}
\def\image{\mathbf{v}}

\maketitle

\begin{abstract}
In this paper, we propose \gan~and~\ganada~-- both incarnations of the state-of-the-art models StyleGan2 and StyleGan2 ADA -- to experiment with their capacity of transfer learning into a rather different domain: creating new illustrations for the vast universe of the game "Magic: The Gathering" cards. This is a challenging task especially due to the variety of elements present in these illustrations, such as humans, creatures, artifacts, and landscapes -- not to mention the plethora of art styles of the images made by various artists throughout the years. 
To solve the task at hand, we introduced a novel dataset, named~\dataset, with thousands of illustration from diverse card types and rich in metadata. The resulting set is a dataset composed by a myriad of both realistic and fantasy-like illustrations.
Although, to investigate effects of diversity we also introduced subsets that contain specific types of concepts, such as forests, islands, faces, and humans. We show that simpler models, such as DCGANs, are not able to learn to generate proper illustrations in any setting. On the other side, we train instances of \gan~using all proposed subsets, being able to generate high quality illustrations. We perform experiments to understand how well pre-trained features from StyleGan2 can be transferred towards the target domain. We show that in well trained models we can find particular instances of noise vector that realistically represent real images from the dataset. Moreover, we provide both quantitative and qualitative studies to support our claims, and that demonstrate that \gan~is the state-of-the-art approach for generating Magic illustrations. Finally, this paper highlights some emerging properties regarding transfer learning in GANs, which is still a somehow under-explored field in generative learning research.

\end{abstract}


\section{Introduction}

\magic~(MTG)\footnote{https://magic.wizards.com/en} is one of the most popular collectible card games in the world. The game is known for its iconic cards, always accompanied by fascinating illustrations. Throughout its almost 30 years of existence, MTG assembled a vast universe of creatures, lands, and artifacts. Each new expansion set comes with different art styles, with 32 possible combinations of colors that alter the design of the cards. This leads to an extensive range of possible gameplays. But it also brings many challenges for generating new illustrations based on a dataset of existing cards. 


In this paper we explore several approaches for training and fine-tuning generative neural networks for generating cards for the game Magic. We begin exploring straightforward strategies which are indeed limited by several aspects of the architecture, and then present options based on transfer learning of large pre-trained GANs. This study shows that it is possible to leverage pre-trained information even when the network was trained in rather distinct domains. For instance, we clearly demonstrate that StyleGAN 2 previously trained to generate human faces can be used to generate a plethora of Magic-related concepts, such as forests, landscapes, human-like characters and the like. Moreover, we present ablative studies carried out in separate subsets of the proposed dataset, allowing us to understand the visual aspects that such networks can easily learn, and those that are harder for them. 

\begin{figure}[!t]
\centering
{\includegraphics[scale=.37]{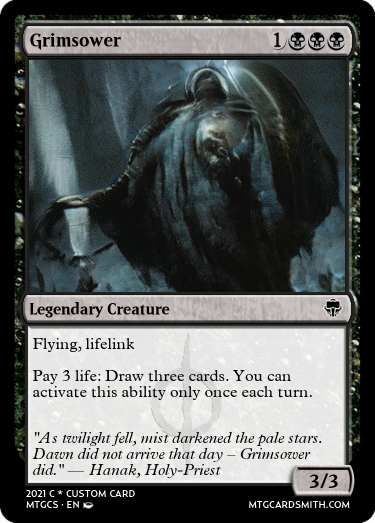}}
{\includegraphics[scale=.37]{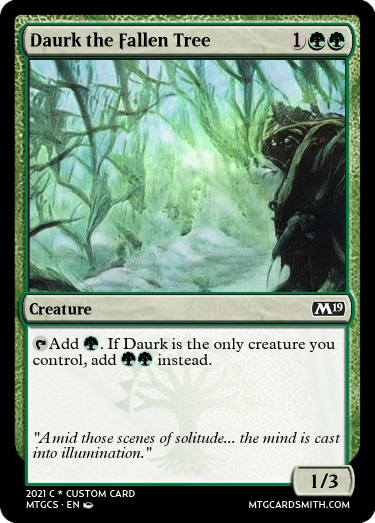}}
{\includegraphics[scale=.37]{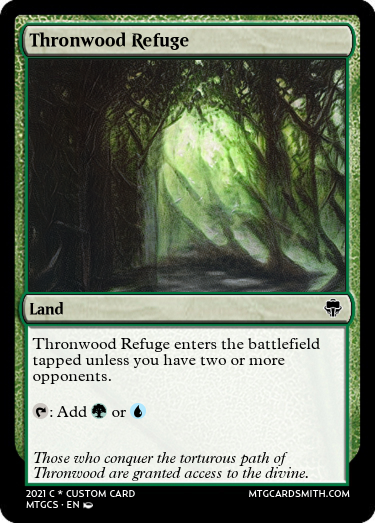}}
{\includegraphics[scale=.37]{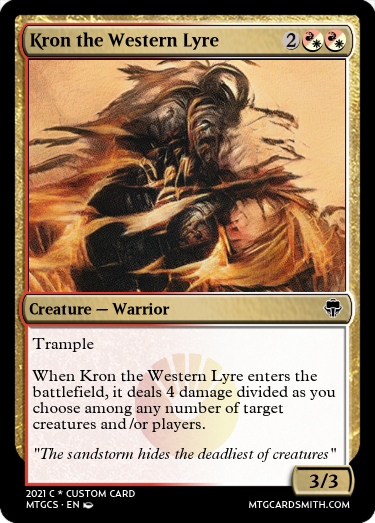}}
\caption{Illustrations generated by the proposed approach. 
Captions of the cards were written by the paper authors.}
\label{fig:examples}
\end{figure}

In summary, we propose a deep learning method, hereby referred as \gan~(see Figure~\ref{fig:examples}), to create new card illustrations for the game \magic~with the aid of Generative adversarial networks (GAN) \cite{goodfellow2014generative}. We introduced a novel dataset composed of more than 60,000 high quality $1024 \times 1024$ card illustrations. Our dataset is labeled with rich metadata, that can be used to identify the original author of the illustration, as well as information regarding category, type of illustration etc. Note that it is complex and computationally costly to generate images in such high dimensionality. To circumvent that we first proposed a DCGAN-based \cite{radford2015unsupervised} baseline that generates lower dimensional images. 


The main contributions of the paper are three-fold: (i) empirical analysis of generative networks for generating artistic looking cards; (ii) extensive evaluation of transfer learning capabilities of state-of-the-art generative models; and (iii) a novel, large dataset containing cards for the game Magic, along to several subsets that can be used to validate transfer learning with few instances in more controlled settings.

\section{Research Problem}

In this paper we propose to handle the problem of card illustration generation for the game \magic. The generation of images for card games is an interesting research problem in many distinct facets. Such illustrations are made out of the time-demanding creative process of multiple artists worldwide, and notably, there is room to improve the artistic capabilities of AI systems. Moreover, it is not a simple task to train a generative model to create such illustrations, given that there is plenty of diversity among the images and the current datasets are limited in number of samples -- not to mention that such images are in high resolutions, e.g., $1024 \times 1024$. GANs tend to generate better results when trained in more specific and controlled domains, such as generation of human faces, horses, cars, beaches, rooms, etc. Therefore, it is unclear whether the current state-of-the-art approaches would be able to represent such large diversity of concepts as those presented in the illustrations. On top of that, one has to recall that the process of training GANs from scratch is often prone to convergence problems and mode collapse. Those can be handled using some best practices, but often large training dataset is also paramount for the model to be able to generalize and generate high-quality high-resolution data.

In addition to experimenting with GANs training from scratch, we aim to uncover some properties of a somewhat unexplored area: transfer learning in GANs. It is widely explored finetuning options applied to visual recognition and language modeling, although there is only few studies on the reuse of pre-trained GANs for different tasks. This paper proposes to answer the following question: \textit{``Do transfer-learning in GANs benefit the generation of artistic illustrations of cards for the game Magic?"}

To the best of our knowledge, this is the first study that handles the task of~\fancymagic~card generation, as well as studying Generative Models transfer-learning capabilities across models trained in completely different domains.

\begin{figure*}[htbp!]
\begin{center}
\includegraphics[width=240pt]{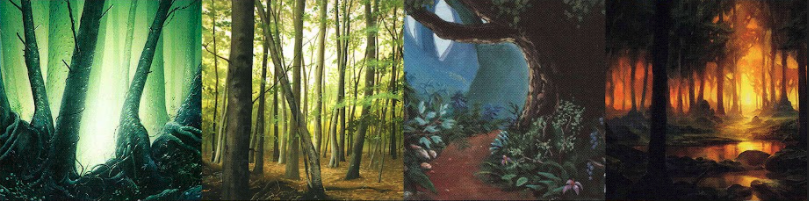}
\includegraphics[width=240pt]{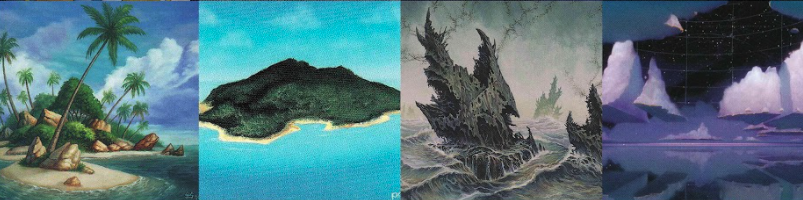}\\
\includegraphics[width=240pt]{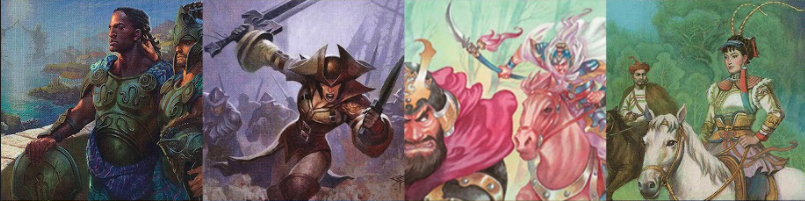}
\includegraphics[width=240pt]{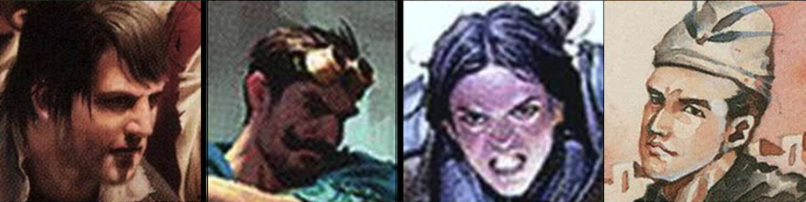}
\caption{Subsets for the \dataset~dataset. Upper-left: samples from \dataset-Forest. Upper-right: samples from \dataset-Islands. Lower-left: samples from \dataset-Humans. Lower-right: samples from \dataset~Faces.}
\label{image_clusters}
\end{center}
\end{figure*}

\section{Building the \gan}
\label{sec:method}


There are plenty of different GAN architectures nowadays, but their transfer learning capabilities are still unclear. Recently, StyleGAN-based architectures did show evidence that they could establish themselves as proper base architectures in order to allow that. This paper helps to shed more light into those questions and try to show their strengths and weaknesses. One should recall that deep neural networks, such as GANs, are data-driven models and require datasets to be trained on.

Such models assume that real images $\mathcal X$ can be generated from vectors $\noise$ randomly sampled from a \textit{priori} distribution $\noisespace{}$ by using a generator model $\mathcal{G}$. In that assumption, we assume that there is a random vector sampled from a normal distribution that can be used as seed to generate a real-looking image, indistinguishable from the real world training data.

Formally, let $\generator$~be the generator function and $\discriminator$~be the discriminator function. Both are incarnated as deep convolutional neural network. $\generator : \noise \mapsto \image$ is the network that produces an artificial image from a random noise $\noise \sim \normal$, where $\normal$ is a normal distribution defined with mean $\mu$ and standard deviation $\sigma$. $\discriminator : \image \mapsto \hat{k}$ is a neural network that tries to identify whether the given image $\image$ is generated or drawn from the real world training data. $k$ is the label that indicates whether the $\image$ is real ($\image_R$) or fake ($\image_F$), so $P(k=Real|k=Fake|\generator(\noise))=D(\generator(\noise))$.

Discriminator has the role of being a neural net supervisor, which is used to allow computing a loss function value $\loss$, and by consequence the estimation of gradients w.r.t. both $\generator$ and $\discriminator$. 

Typically, the loss value of the generator would be the opposite of the discriminator loss, i.e., $\loss_G = -\loss_D$, which \textit{per se} defines an instance of Adversarial Training~\cite{goodfellow2014generative}. 

In this setting, we optimize the discriminator weights $\theta_d$ by minimizing the binary cross entropy of the predictions for both real and generated images

\begin{equation}
\Delta \theta_d \frac{1}{m} \sum^{m}_{i=1} \Big [ \log \discriminator(\image_i) + \log (1-\discriminator (\generator (\noise_i))) \Big]
\end{equation}

where $m$ is the number of instances in the mini-batch, and $\image_i$ is the $i^{th}$ image drawn from the real data distribution $\imagedist{}$, and $\noise_i$ is the noise sampled from $\noisespace{}$ for that iteration. For optimizing the weights $\theta_g$ of the synthesis network, we use the opposite of the loss function for the $\discriminator$ as shown in Equation~\ref{eq:discriminator_loss}.

\begin{equation}
\label{eq:discriminator_loss}
\Delta \theta_g \frac{1}{m} \sum^{m}_{i=1} \Big [ \log (1-\discriminator(\generator(\noise_i))) \Big ]
\end{equation}

The overall optimization problem objective is then formulated as the following  $\min \max$ game,

\begin{equation}
    \min_\discriminator \max_\generator 
    \mathbb{E}_{\image \sim \imagedist} 
        [\log(\discriminator (\image))] 
    + \mathbb{E}_{\noise \sim \mathcal Z}
        [\log (1-\discriminator(\generator(\noise)))]
\end{equation}

where the synthesis net is forced to generate images more alike to the real data distribution to fool the discriminator net -- and at the same time, the discriminator has to improve to be able to distinguish between fake $\image_F$ and real $\image_R$ images. 

Such training is much more unstable than supervised ones, and require a fine balance between the learning capability of both $\discriminator$ and $\generator$. There has been extensive work on improving GAN training stability, for instance Spectral Normalization, SAGAN, DCGAN, ProGAN and StyleGAN.

\subsection{\gan}

In recent architectures, such as the StyleGAN, the noise vector $\noise{} \in \noisespace{}$ is first projected via a mapping network $f$ to an intermediate latent code $\projnoise \in \projnoisespace$, i.e., $f: \noisespace \mapsto \projnoisespace$. In this strategy, instead of simply using the sampled noise, we first project it using a non-linear multi-layered perceptron $f$. The projected noise vector $\mathbf w$ is used in multiple layers to control the synthesis network $\mathcal G$. The approach used to condition $\mathcal G$ to the projected noise is via Adaptive Instance Normalization (AdaIN). Such an approuch has been widely used to allow style transfer networks~\cite{karras2019style} -- giving the StyleGAN its name.

Such an approach has been used for solving many different tasks, but most notably it has been widely used to allow style transfer networks~\cite{karras2019style} -- giving the StyleGAN its name.

By using such affine transformations to control the synthesis network the intermediate space $\mathcal W$ becomes less entangled than when using directly the input noise vector $\mathbf{z}$, which adds more stochastic variation to the network. 

The subsequent version of StyleGAN, namely StyleGAN2~\cite{karras2020analyzing}, brings some improvements to its predecessor such as the path length regularizer, used along to the generator to make it easier to invert. This is specially important when we want to optimize the specific noise vector $\noise$ that represents a given real or generated image $\image$. StyleGAN2 also introduces larger networks, which with the aid of Progressive Growing can stably be optimized to generate high-resolution images with sizes up to $1024 \times 1024$.

StyleGAN2 model has been shown by many researchers as a powerful tool for unconditional image generation \cite{back2021finetuning, detached_stylegan2, vavilala2021controlled, abdal2019image2stylegan, blood_cells_stylegan2}, 
and it still matched its successor StyleGan3 in FID comparison \cite{karras2021aliasfree}.
We chose to adopt such architecture as the base for this study considering the generated images quality. Therefore, we leverage available pre-trained networks for evaluating the StyleGAN 2 transfer learning capability. 

Considering that StyleGAN2 promotes disentangled and hierarchical noise introduction to the network, allied to the inversion improvements, we believe that it is possible to finetune such model so as to smoothly adjust the features learned by the model into the Magic illustrations domain. Models tuned in this strategy are called~\gan.

\subsection{\ganada}

Notably, there is a rather limited amount of Magic card illustrations. Thus, in the context of limited data, the StyleGan2 model may suffer from discriminator overfitting, causing the feedback to the generator to become meaningless and the training to diverge. One of the most common alternatives to prevent overfitting is to adopt stronger data augmentation schemes. Note that in GANs, augmentation practices may affect the generated images, causing \textit{transformation leaking} or making the discriminator gradient useless.

To circumvent those issues we employ the strategy presented by StyleGan2-ADA~\cite{karras2020training}~to train \gan-ADA models. They use a wide range of invertible augmentations applied to both nets in an adaptive way. Invertible transformations have been shown to be non-leaking and by using them in both fake and real images the discriminator will not be able to simply overfit the augmentation style to discern between them. Such improvements may as well benefit transfer-learning capabilities in small dataset regimes, which is the case for generating Magic illustrations.


\subsection{Transfer learning}

One of the main goals of this paper is to understand whether pretrained StyleGAN-based ease the process of generating illustrations for the Magic cards. We expect them to be able to do a smooth domain shift towards the target application. To observe this effect, we have to register the progression of the generated images during each epoch of the finetuning phase. A smooth domain shift would be characterized by a sequential morphing from the original generated images towards the target domain. Note that if the training diverges, or the pretrained features are not useful, the network would most likely \textit{forget} its original patterns, to learn new ones from the target data.

We suspect that by using the previous knowledge acquired from the pre-training phase as base to shift towards a new target domain will reduce the need for large datasets to achieve high quality image synthesis. In addition, it is likely the case that we would achieve significantly better results than training from scratch.

In our experiments, we proposed a variety of transfer learning scenarios to understand the impact of the pre-training data distribution. For instance, we uncover the impact of finetuning models that had source domains similar or very distinct to the target dataset. We also test cases where source and target datasets have similar or very distinct levels of diversity. Results of those experiments can be seen in Section~\ref{sec:experimental}.


\section{\dataset~Dataset}

In this section we detail: (i) a preliminary analysis using a publicly available~\magic~card illustration dataset to uncover its limitations and understand the complexity of the task at hand; and (ii) the creation of a novel dataset, namely \dataset, that is intended to fix all problems faced in the preliminary analysis, and at the same time allow for great many future research uses.

\subsection{Preliminary Analysis}\label{preliminary_analysis}

For training models that can generate Magic card illustrations it is necessary to have a dataset with real illustrations at hand. To the best of our knowledge there is a public dataset, referred as $Magic 1024 \times 1024$\footnote{https://www.kaggle.com/joaopedromattos/magic-the-gathering-cards-1024x1024}, that contains 49 thousand images from different card illustrations. We did preliminary experiments using a sample from that dataset, and all models failed to properly approximate the real data distribution. That dataset contains a diverse set of cards, but no information regarding card types and card content. Therefore, making it an overly complex function for the models to learn.

\subsection{Creation of the \dataset~dataset}


In order to fix the previously mentioned issues, we gathered a novel dataset that we hereby call \dataset. It has more card illustrations samples ($62,000$ images versus $49,000$), and also richer in metadata that allows us to precisely create subsets for specific concepts.

\begin{figure*} [htbp]
\begin{center}
\includegraphics[width=\textwidth]{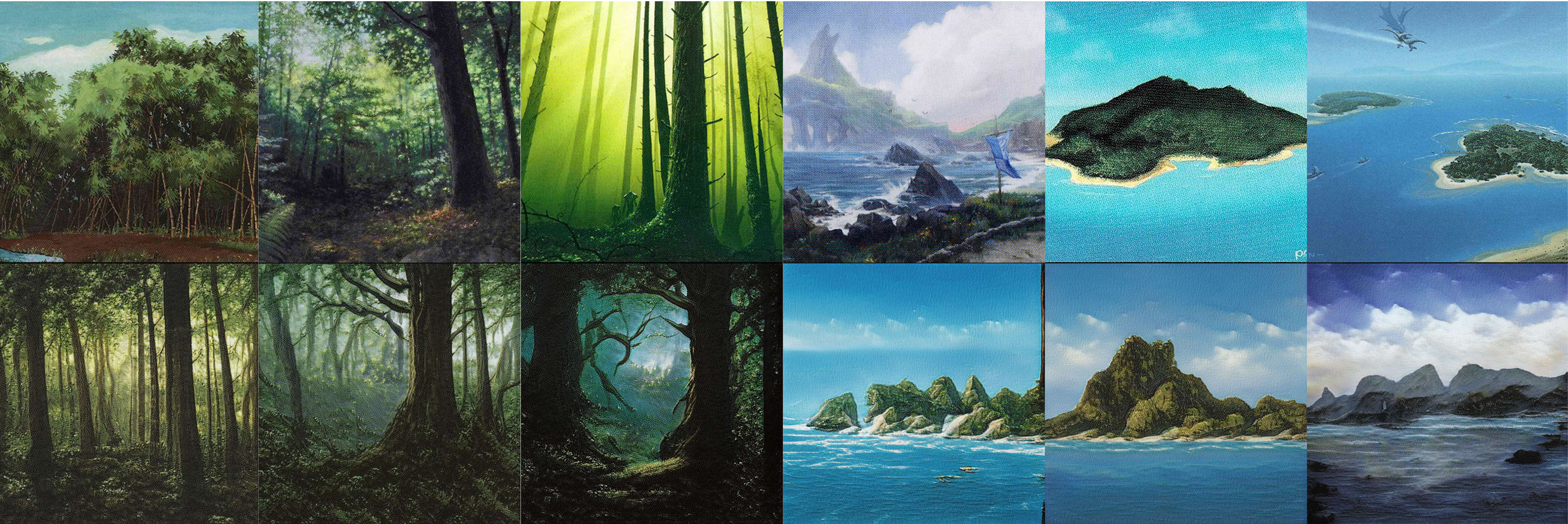}
\caption{Comparison of training results for the "MTG-Island" (left) and "MTG-Forest" (right) dataset with MagicStyleGAN-ADA. The first line shows samples of the source datasets, while the second line presents curated examples of generated illustrations.}
\label{curated_islfor_sg2ada}
\end{center}
\end{figure*}

We built the \dataset~dataset by downloading card images from the Scryfall\footnote{https://scryfall.com/} website. Then, we upscaled lower resolution images and extracted $1024 \times 1024$-sized center crops from each illustration. Hence, we standarize all images to have the same resolution. Our dataset contains a multitude of card types (creatures, lands, artifacts) and colors (white, blue, black, red, green, and its combinations). The collected metadata dataframe contains 82 columns, such as: keywords, name of the card, oracle text, produced mana, color indicator, and many more. Generation of card illustration is one but many other possible tasks that can be handled using this dataset, such as: (i) text-to-image generation; (ii) card name prediction; (iii) modeling of oracle text; (iv) card authorship identification; and (v) keyword generation. 

Focusing on the goal of this study, we created five main subsets out of \dataset, namely Cards, Islands, Forest, Humans, and Faces. Those sets would contain only illustrations related to their categories, which helps reducing concept diversity, and may allow for training generative networks for each card type. Following, we describe all subsets: \dataset-Cards~is the complete dataset, otherwise referred simply as \dataset; \dataset-Islands has illustrations of “basic land” type cards named “Island”; \dataset-Forest has images of “basic land” type cards called “Forest”; \dataset-Humans contains a reduced set of human-like creatures; and \dataset-Faces comprises manually extracted crops from the \dataset-Humans illustrations focusing only on the character faces. See~Table~\ref{tab_dataset} for more details and statistics on the introduced subsets, and also some related datasets for scale comparison. 

\begin{table}[htbp]
\caption{Information on each dataset that was used for training and experiments.}
\centering
\begin{tabular}{lcc}
\toprule
Dataset & Image Size & Samples \\
\midrule
Flickr-Faces-HQ (FFHQ) &  $1024\times 1024$ & 70,000 \\
AFHQ Dogs & $1024\times 1024$ & 5,000 \\
MetFaces & $1024\times 1024$ & 1,336 \\
Magic 1024$\times$1024&  $1024\times 1024$ & 49,000 \\
\midrule
MTG-Cards &  $1024\times 1024$ & 62,000 \\
MTG-Islands &  $1024\times 1024$ & 614 \\
MTG-Forests &  $1024\times 1024$ & 625 \\
MTG-Islands+Forests &  $1024\times 1024$ & 1,239 \\
MTG-Humans &  $1024\times 1024$ & 1,376 \\
MTG-Faces &  $1024\times 1024$ & 1,161 \\
\bottomrule
\end{tabular}
\label{tab_dataset}
\end{table}

Note that we did build \dataset-Humans by collecting creature cards of two compound types: “human warrior” and “human soldier”. This creates a subset that share important main characteristics though with reduced variety in style. Illustrations from that set often present similar clothing such as full-body armour, etc.

We formed \dataset-Faces by manually extracting faces of human creatures. We also filtered some outlier samples that differed too much from the expected human face (even for an illustration) or that had a resolution that would hinder the interpretation of the image.

It's worth noticing that these datasets, although more constrained, still have great deal of variety in design and style from the various artists that created their illustrations throughout the years. 

We also include some experiments to understand the diversity impact in training such models, in which we train \ganada~conjointly in multiple sets, such as the \dataset-Islands+Forests split that combines both \dataset-Islands and~\dataset-Forests samples. 

\section{Experimental Setup}

In this section we explain the methodology for running all experiments, including baseline models, evaluation procedures and metrics, as well as the hyper-parameters used. 

\subsection{Baseline models}

For the experiments listed in this paper, we trained GANs both from scratch and applying transfer learning on pre-trained models.
In order to validate our hypotheses regarding the behavior and advantages of \gan~and~\gan-ADA we use  DCGAN and StyleGAN trained from scratch as main baselines. These baselines can help us to understand if it is advantageous to perform finetuning, and if ADA-related models do perform better in limited data regimes. 

For running the transfer learning experiments we used the StyleGan2 and StyleGan2 ADA pre-trained models made available by NVIDIA in their public Github repositories. Code was written in the TensorFlow framework. For training the \gan~model, we used the StyleGAN pre-trained with the FFHQ dataset as the backbone of our experiments. Training procedures with \gan-ADA used a series of other trained models: FFHQ, AFHQ Dogs and MetFaces. By finetuning models pre-trained in different domains (faces, dogs, etc) we can understand the effects that the origin domain has in the transferability of the learned features towards the target domain. In addition, we can verify the truthfulness in the assumption that models pretrained in domains similar to the target ones do transfer better and allow for improved generation quality.


\subsection{Hyper-parameters}

All of our models were trained in single GPUs, with a batch size of 4, though with gradient accumulation for 8 steps to get gradients from 32 instances. We made use of NVIDIA Tesla P100 and V100 GPUs. Complete finetuning procedure takes 24 hours in a single P100 GPU, while the V100 is almost three-fold faster. We estimate that all experiments took more than 480 GPU hours to run. All models were optimized with Adam with $\beta1=0.0,\beta2=0.99$ and learning rate of $2 \times 10^{-3}$ for both discriminator and generator networks. For data augmentation sampled images were randomly mirrored left to right, but not top to bottom. Augmentation hyperparameters used for \gan-ADA~were pixelblitting and geometric transformations. Generated images had truncation at 0.7.

\begin{figure*} [htbp]
\begin{center}
\includegraphics[width=250pt]{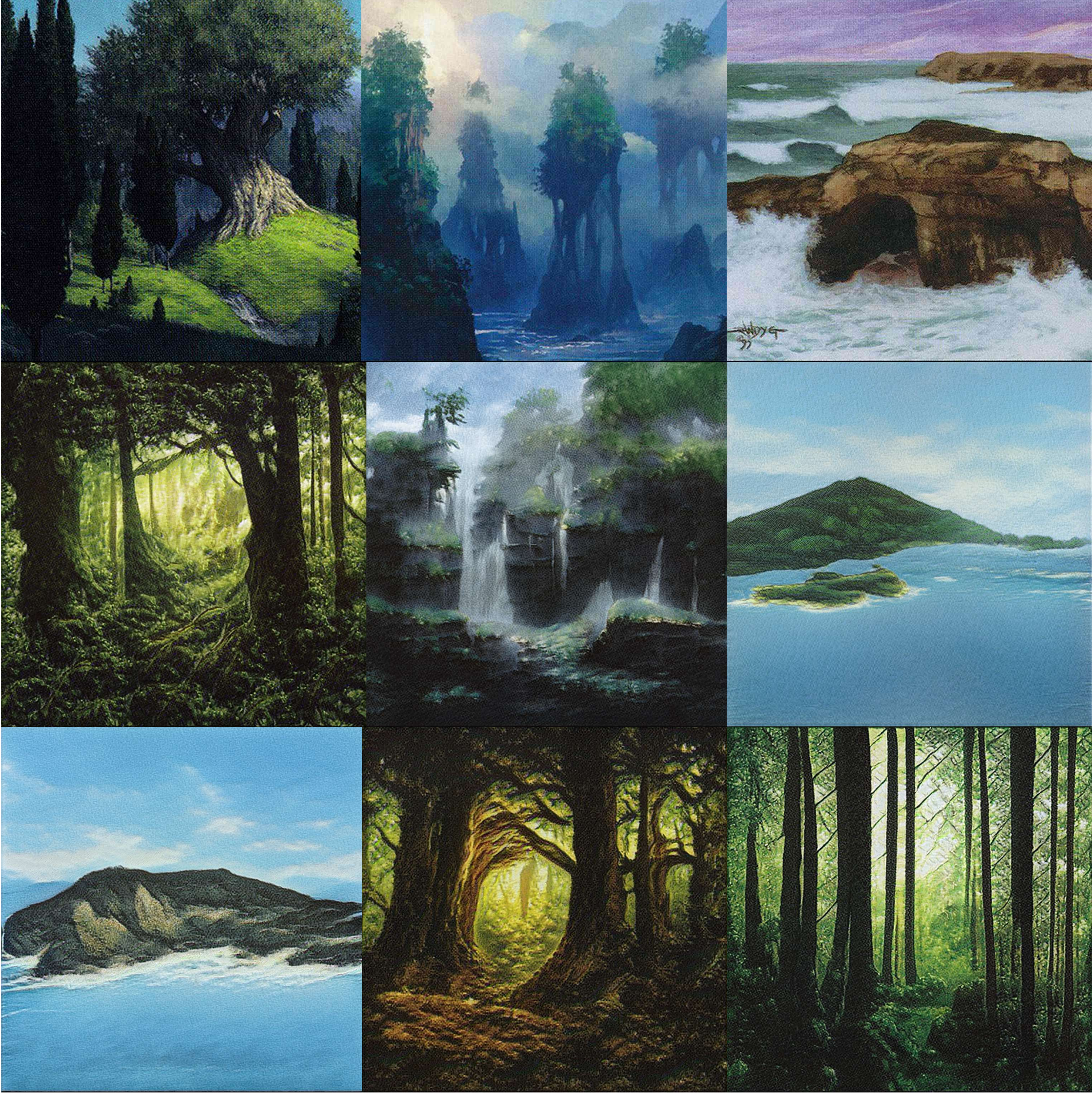}
\includegraphics[width=250pt]{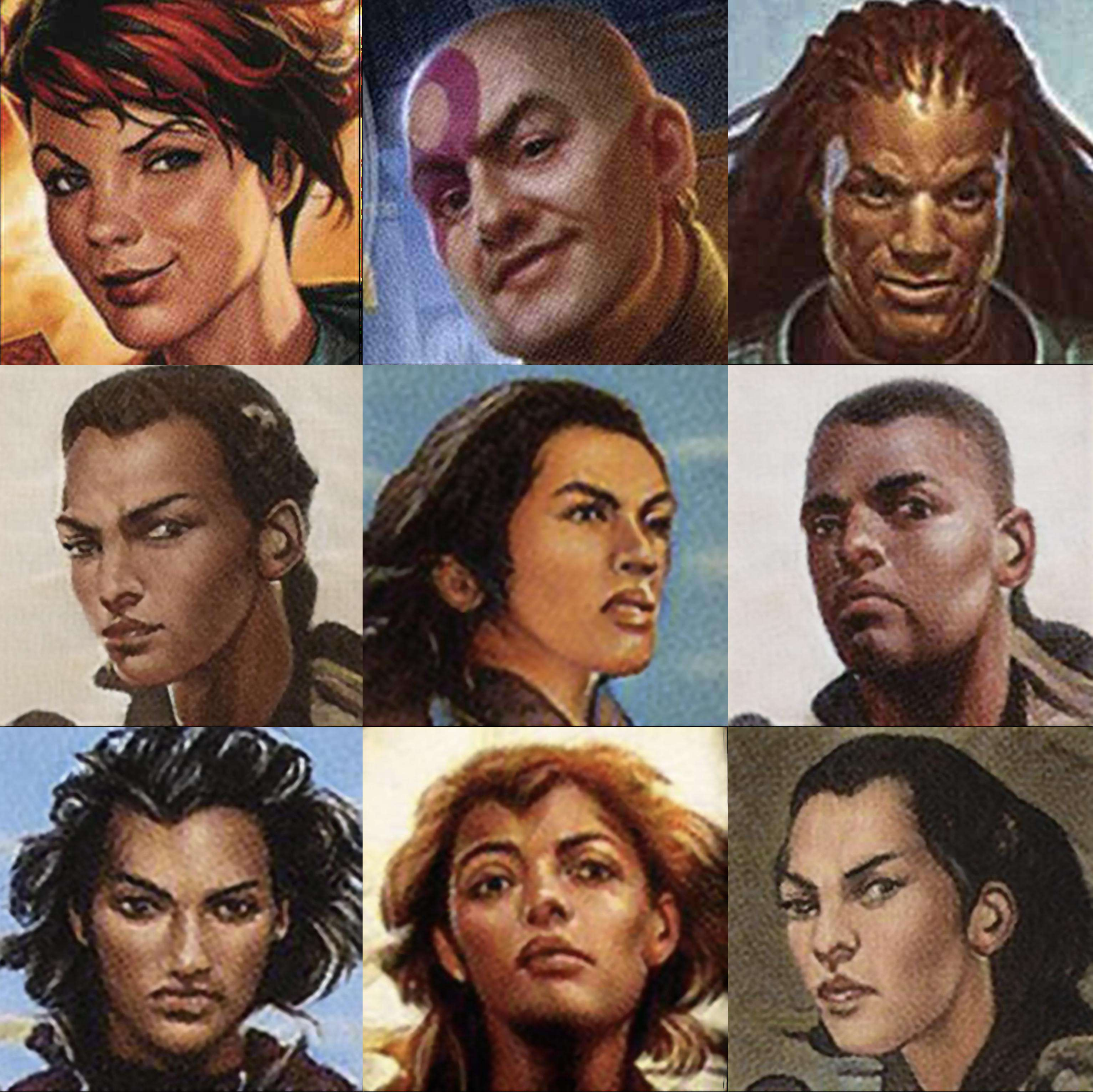}
\caption{Left: Curated results from \ganada~ trained with MTG-Islands+Forests dataset. Top line shows samples of the target dataset, and generated images are shown on the middle and bottom line. Right: Curated results from \ganada~ trained with MTG-Faces dataset and FFHQ dataset as source domain. Post-processing with denoise. First line shows samples of the "MTG-Faces" dataset, and the second and last line shows examples of generated images.}
\label{curated_i_and_f_sg2ada}
\end{center}
\end{figure*}

\subsection{Model evaluation}

This study employs two largely used metrics for quantitative evaluation of GANs, specifically, FID and KID. 

FID (Fréchet Inception Distance) is a well accepted metric for the context of GAN performance \cite{heusel2018gans}. It uses the InceptionV3 model with the goal to capture the similarity of generated images and real ones. FID equation in Equation~\eqref{fid_formula} is a distance measurement between the distributions of the features from images generated when compared to features of real samples from the dataset. Low metric values represent high similarity, and therefore, lower \fid values are better as indicated by the down arrow. FID values in this work were calculated for 10,000 generated images.

\begin{equation}
\label{fid_formula}
 FID = ||\mu_r – \mu_g||^2_2 + Tr(  \Sigma_r + \Sigma_g – 2\sqrt{\Sigma_r \Sigma_g})
\end{equation}

Kernel Inception Distance (KID) is somewhat similar to FID, and describes the squared MMD (Maximum Mean Discrepancy) between Inception representations \cite{binkowski2021demystifying}. KID uses the polynomial kernel, defined in Equation~\eqref{kid_kernel}, to avoid correlations with the objective of GANs and to avoid tuning any kernel parameter. KID estimates are unbiased, and standard deviations shrink quickly even for small datasets.

\begin{equation}
\label{kid_kernel}
 k(x,y) = \Big{(}\frac{1}{d}x^{\intercal}y + 1 \Big{)}^3 
\end{equation}

It has been mentioned that FID is not an ideal metric when dealing with limited data \cite{karras2020training}, because the estimator of FID is biased. Therefore, we opted to also report KID values since it is more descriptive in practice for the aforementioned reasons. Similar to FID, the lower the \kid~value, the better.

\section{Experimental analysis}
\label{sec:experimental}

The experiments tested the generation of illustrations in a number of datasets with large aesthetics and pattern variation between them. In this section we discuss results achieved by~\gan~and~\ganada, that leveraged transfer learning from pre-trained StyleGan2 and StyleGan2 ADA models. 

\begin{table*}[htbp]
\caption{Quantitative results.}
\centering
\begin{tabular*}{\textwidth}
{@{\extracolsep{\fill}}lrrrrrr}
\toprule
 &   \gan &  \ganada &   DCGAN &  \gan &   \ganada &   DCGAN \\
Dataset & \multicolumn{3}{c}{\fid} & \multicolumn{3}{c}{\kid ~$\times10^2$} \\
\midrule
 MTG-Cards           &      \textbf{0.678} &                   0.803 &                      0.792 &              \textbf{2.693} &           4.043 &           22.820 \\
 MTG-Islands         &       \textbf{0.986} &                   1.201 &                       29.743 &            \textbf{1.457} &           3.581 &           700.400 \\
 MTG-Forests         &        \textbf{1.250} &                   1.547 &                       6.237 &              \textbf{2.243} &           4.581 &           142.140 \\
 MTG-Islands+Forests &      \textbf{0.845} &                   1.026 &                       20.341 &            \textbf{3.115} &           3.505 &           230.830 \\
 MTG-Humans          &                   1.141 &        \textbf{1.035} &                       8.331 &                       5.442 &  \textbf{3.080} &           264.280 \\
 MTG-Faces           &       \textbf{0.747} &                   0.840 &                       1.649 &                \textbf{2.327}&           4.872 &           31.880 \\
\bottomrule
\end{tabular*}
\label{tab_quantitative_results}
\end{table*}

\subsection{Transfer learning yields far better results}\label{tl_raw_mtg_cards}

For the main results of the paper we used the FFHQ pre-trained StyleGan2 model as the source domain dataset. Interestingly, for most of the experiments FFHQ images are quite different: they differ in texture, proportions and semantics. 

\begin{figure} [htbp]
\begin{center}
\includegraphics[width=240pt]{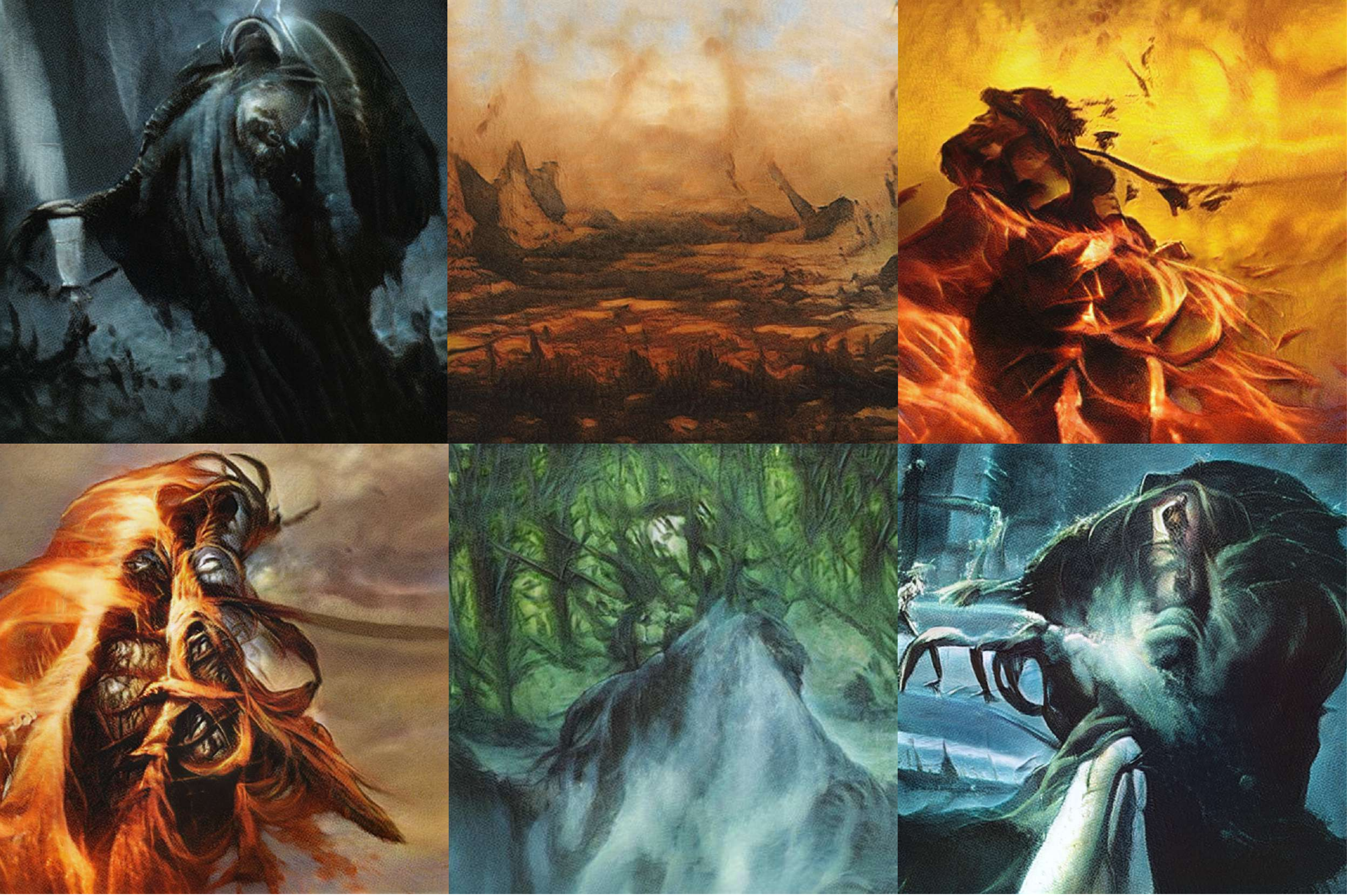}
\caption{Curated images generated by~\gan: the model was fine-tuned in a sample of 10 thousand images of "Magic 1024$\times$1024".}
\label{fig:dataset_samples_mtg_curated_sg2}
\end{center}
\end{figure}

Figure~\ref{fig:dataset_samples_mtg_curated_sg2}~showcases curated output from the trained model on \dataset-Cards. When trained in MTG-Cards, our approach was able to produce good results despite the large diversity in the training data. It performed significantly better than DCGAN in terms of KID (Table~\ref{tab_quantitative_results}). Nonetheless, most of the generated images did not have the potential to be mistaken by real MTG illustrations. This shows that for better results, one should consider training separate models for each card type.

\subsection{Subsets allow for better generation}

After training with~\dataset-Cards, the next experiments tested more restricted datasets by using the created subsets: Islands, Forests, Humans, and Faces. Since these datasets have a small number of samples, we opted for the reporting StyleGan2 ADA architecture first. The curated output from the generated islands and forests can be observed in Figure~\ref{curated_islfor_sg2ada}. The training is successful in transferring the knowledge from the FFHQ source dataset to the target domains, even when the domains are very different. The generated images are far better looking and more consistent than the results from section \ref{tl_raw_mtg_cards}, which suggests that creating per-concept subsets positively affect the result of the model training. 

\begin{figure*} [bpht]
\begin{center}
\includegraphics[width=380pt]{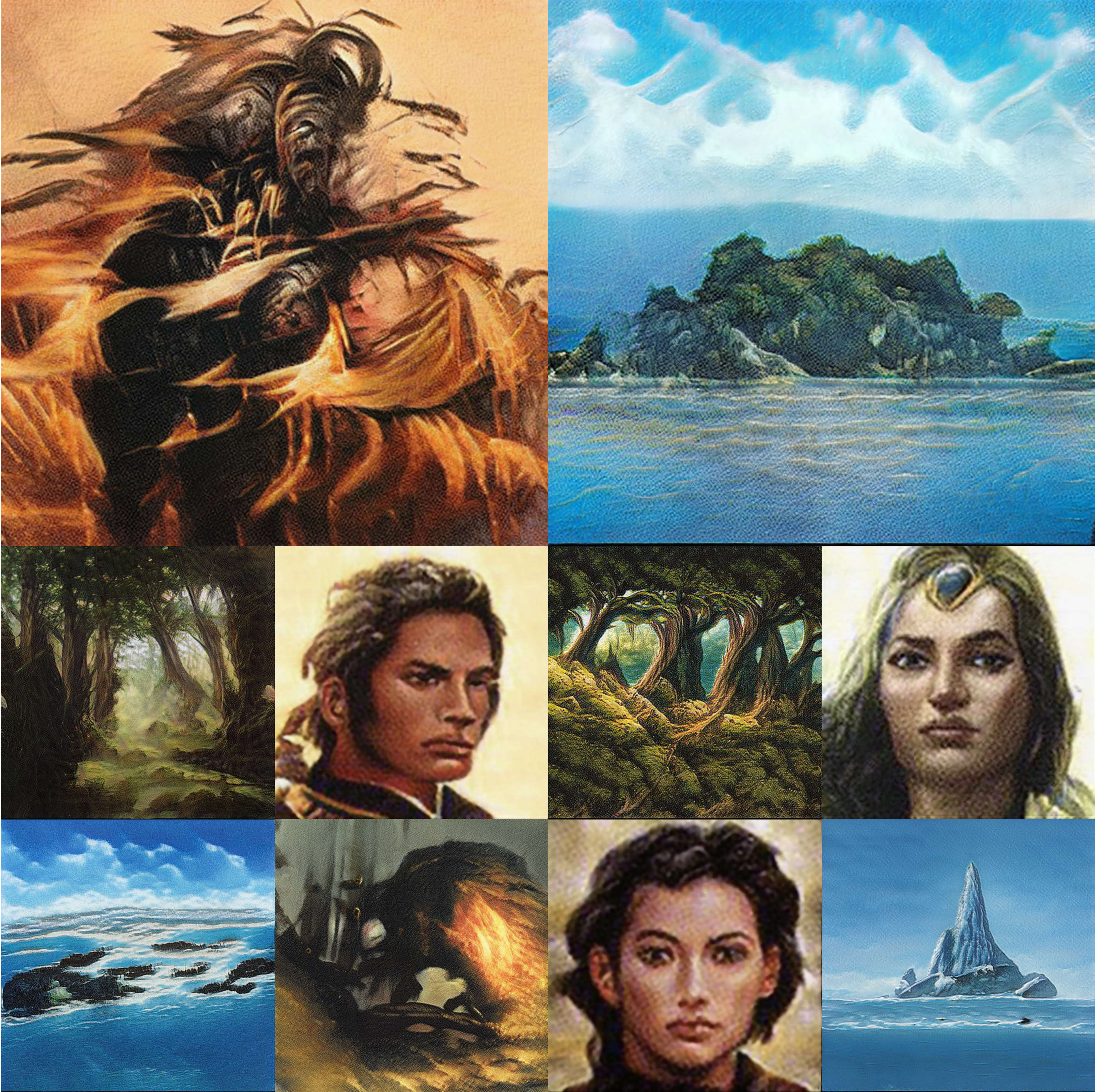}
\caption{Examples of illustrations for \magic~from the models \gan~and\ganada, models proposed throughout the paper, using the dataset \dataset~and the subsets created in this work.}
\label{fig:results_methods_generation}
\end{center}
\end{figure*}

Furthermore, we ran an experiment that included both images from a combined version of MTG-Islands and MTG-Forests, that is the MTG-Islands+Forests. Figure~\ref{curated_i_and_f_sg2ada}~presents curated generated images from \ganada~model trained in that data set. The model was still able to learn differences between the two concepts, and was able to generate both types of classes. That being said, the model was less consistent in generating high quality samples. Higher resolution images from models trained in all subsets can be seen in Figure~\ref{fig:results_methods_generation}.

\begin{figure*}[ht!]
\begin{center}
\includegraphics[width=\textwidth]{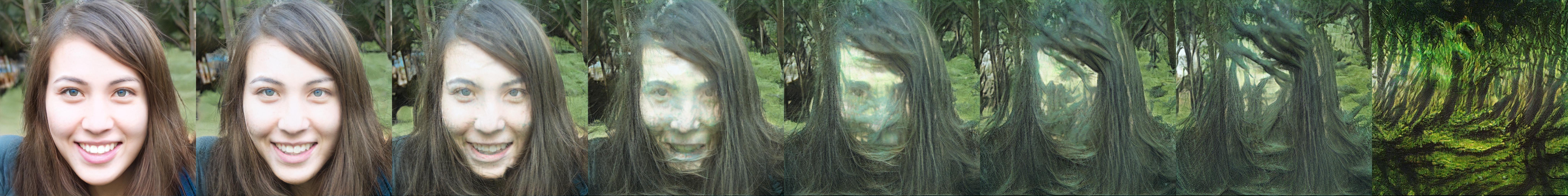}
\includegraphics[width=\textwidth]{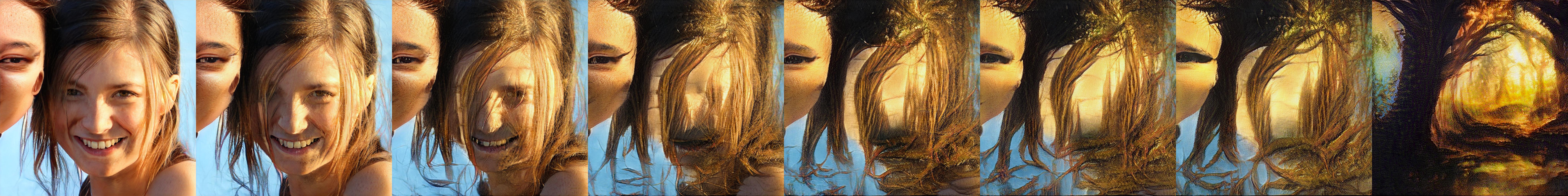}
\includegraphics[width=\textwidth]{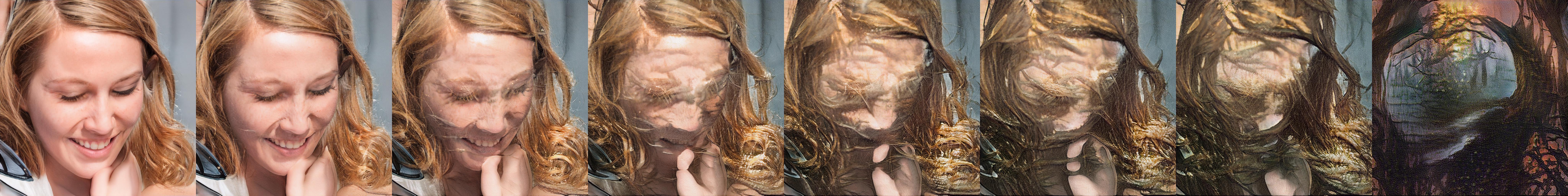}
\caption{Per epoch evolution of the generated image for the model~\gan~ trained with \dataset-Forest dataset. First to penultimate columns are from subsequent training epochs. The last one showcases the image generated by the end of the training procedure. Note: image was compressed due to maximum paper size policy for the submission.}
\label{fig:evolution_training}
\end{center}
\end{figure*}

\subsection{Variety on target domain impacts generation quality}

Perhaps the more challenging subset from~\dataset~is the~\dataset-Humans. That is possibly the case given that humans do present themselves as complex forms composed by a variety of body parts, poses, movements, and clothing. Experiments ran in this subset demonstrated that \gan~was able to learn some high level human features, though failed to draw them in detail and in a consistent fashion. 

The network is able to understand the basic structure of humans to a certain point, such as the position of the head and the presence of eyes. \ganada~was also able to generate the texture of armors (widely present in "MTG-Humans"). Though, all models failed to generate realistic samples, since  this dataset is incredibly complex as there is substantial diversity among the dataset samples -- specially when compared to the number of instances. 

In contrast, \gan~and~\ganada~were able to quickly learn to generate proper samples for the "MTG-Faces" dataset, which is focused on  generating new illustrations of character faces. Figure~\ref{curated_i_and_f_sg2ada}~shows faces generated by \ganada. We observed that the manual crop of each face significantly helped the learning process of the network. That is expected once the face pattern becomes easier for the network to understand and to related to the origin features. 

\subsection{Effects of the base model domain}

In this experiment we explore the relation between the source domain dataset and the results of the generated images. It has been pointed out that transfer learning success seems to depend primarily on the diversity of the source dataset, instead of similarity between subjects \cite{karras2020training}. 

The results displayed in Table~\ref{source_domain_fid}~show that the similarity does have an important effect on transfer learning. For MTG-Faces, the best results were achieved with the source domains of both faces in paintings and high quality faces pictures. Notably, the best results for MTG-Islands+Forests~results are from models finetuned from FFHQ pretrained weights, which indicates that well trained models can provide useful features to a plethora of domains. Nonetheless, AFHQ Dogs also provided competitive results. Such results help demonstrating that transfer learning could be much more further explored in generative models. 

\begin{table}[htbp]
\caption{Effects of the source domain on training~\ganada.}
\begin{center}
\begin{tabular}{lccc}
\toprule
 & FFHQ & AFHQ Dogs & MetFaces \\
 Dataset & \multicolumn{3}{c}{\fid} \\
\hline
MTG-Faces  & 0.840 & 1.038 & 0.804  \\
MTG-Islands+Forests  & 1.026 & 1.111 &  1.463 \\
\bottomrule
\end{tabular}
\label{source_domain_fid}
\end{center}
\end{table}


\section{Quantitative Results}

\begin{figure} [htbp]
\begin{center}
\includegraphics[width=\linewidth]{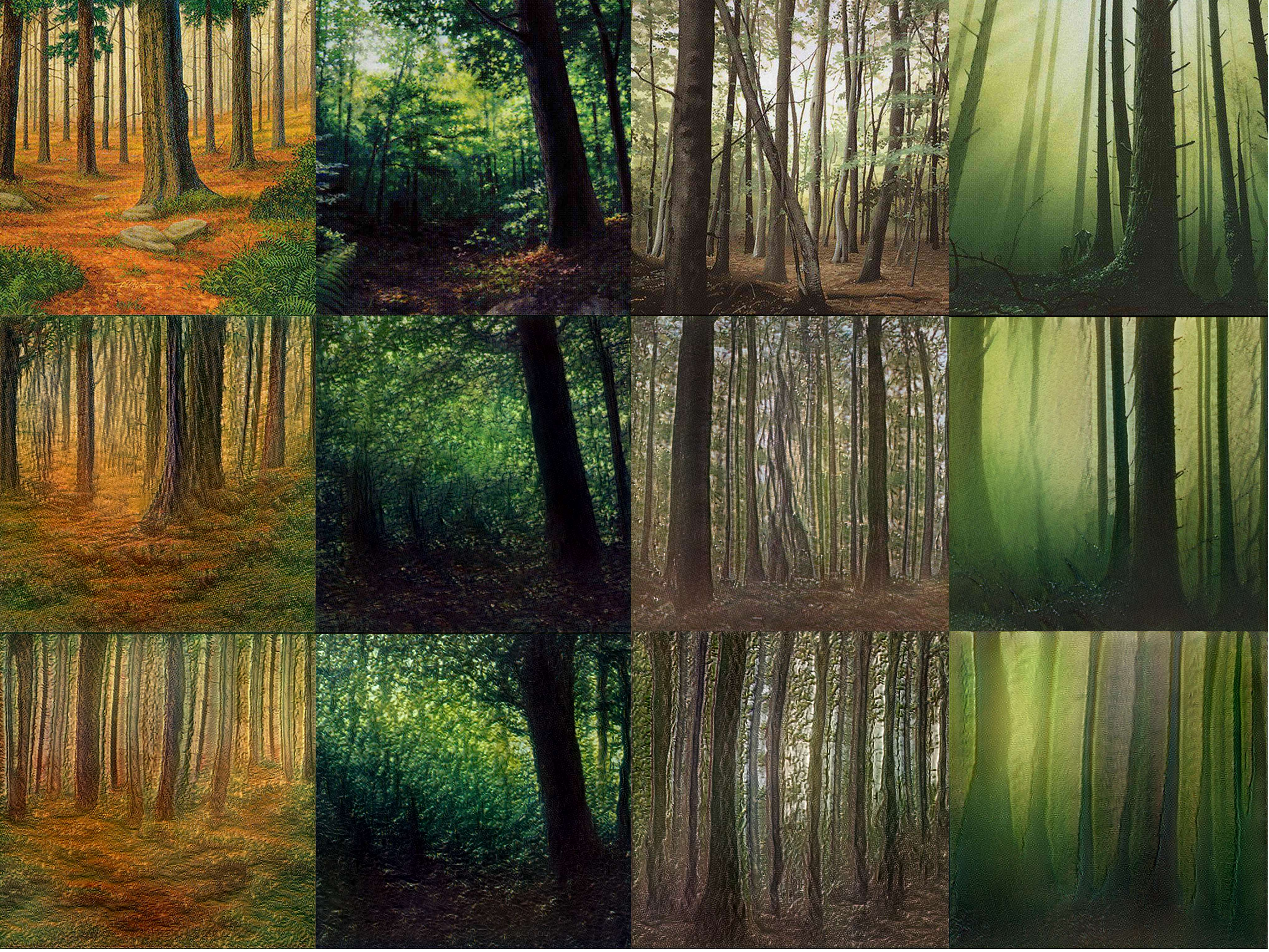}
\caption{Comparison of the projection of target images from MTG-Forests on the latent space of \ganada~ trained with MTG-Forests using transfer learning with the weights from FFHQ and StyleGAN2-ADA trained without transfer learning with MTG-Forests. Target images on the top, \ganada~ in the middle row and StyleGAN2 on the bottom.}
\label{fig:noise_scratch_comparison}
\end{center}
\end{figure}

We first report the quantitative metrics for our models, and a DCGAN baseline, considering all datasets. Table~\ref{tab_quantitative_results}~depicts that \gan~and~\ganada~presented very similar performance, and both greatly outperformed the baseline trained from scratch. It is important to mention that some results can be counter-intuitive. For instance, StyleGan2-ADA trained from scratch using \dataset-Forest got a FID of $1.498$ and KID of $0.039$, which is similar to the transfer learning results with the same setup. That being said, we noticed that the transfer learning model generates reasonable results much faster. In addition, experiments shown in~Figure~\ref{fig:noise_scratch_comparison}~clearly demonstrate that the projection of target data from \dataset-Forest on the latent space of StyleGAN2-ADA and~\ganada~shows a big difference in quality, with the latter getting the upper hand. 

Relative low scores for \dataset-Cards and ~\dataset-Humans were achieved on all GANs while a manual evaluation of the generated images points to a different conclusion, as we discuss in Section~\ref{sec:noise_vectors}. FID and KID rely heavily on InceptionV3 model \cite{szegedy2015rethinking}, which loads weights pre-trained on ImageNet \cite{imagenet2009}. Therefore, it is unclear how well these metrics work for a domain such as \magic, that is rather distinct from the ImageNet-related ones. We aim to address this considerations in future works, maybe proposing a more reliable metric considering newer models, such as CLIP~\cite{radford2021learning}.

\section{Visualizing the learned representations}

In this section we analyze the trained models to evaluate their representation learning capacity. First, we validate our assumption that when finetuning StyleGAN-like architectures we can see a clear domain shift as the network leverages human-face like features to generate the target domain shapes. Then we evaluate the quality of the \textit{prior} distribution when we try to find the noise vector $\noise$~that when fed to the synthesis network ($\generator$) generates the real world image.

\subsection{Visualizing the representation shift}

Figure~\ref{fig:evolution_training}~clearly shows that when using a model pre-trained to generate high-quality faces as base, the network learns to adapt such features as to generate shapes, colors and textures from the target domain. It is amusing to see that during the initial stages of the optimization it makes a mashup of both origin and target domains. As the training continues, the network forgets about face features and concerns more in generating detailed card illustrations.

It is relevant to notice that even for very distinct target domains, training \gan~and~\ganada~ could in fact leverage from transfer learning based on StyleGan2 models. Our models were able to quickly understand the visual aspects of the proposed datasets and to generate possible illustrations for the game \magic.

\subsection{Finding the noise vector for real images}\label{sec:noise_vectors}

\begin{figure*} [htbp]
\begin{center}
\includegraphics[width=\linewidth]{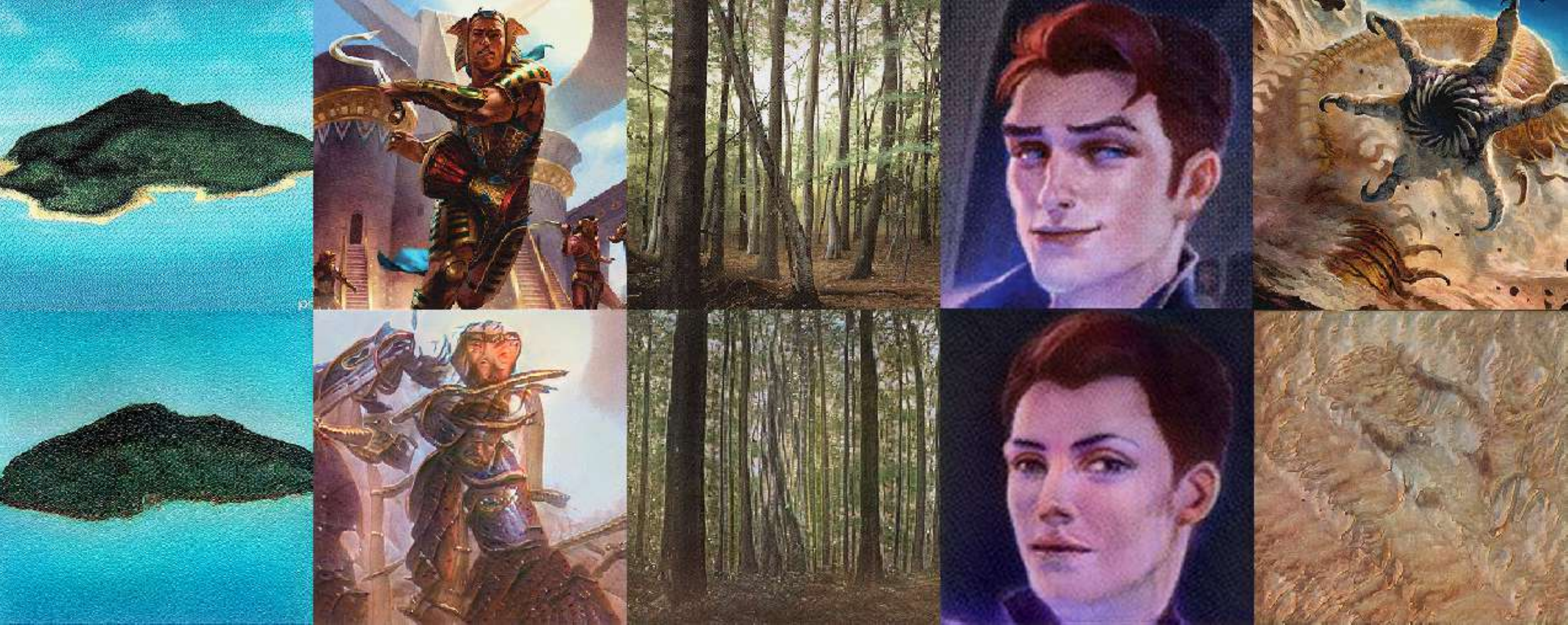}
\caption{Target images and their respective projections on \ganada's latent space. Top line shows the real world target images, and the at bottom line we depict their respective projections. From left to right: samples from MTG-Islands, MTG-Humans, MTG-Forests, MTG-Faces and MTG-Cards, respectively.}
\label{fig:real_images_projected_on_latent_space}
\end{center}
\end{figure*}

Figure~\ref{fig:real_images_projected_on_latent_space}~shows the projection of target images onto latent space of \ganada. I.e., we optimize the noise vector $\noise$ that generates the closest image of the given \textit{real} image. By analyzing the projection, it becomes clear that the network is able to generate better approximations when dealing with more uniform datasets like MTG-Forests, MTG-Islands and MTG-Faces. On the other hand, for models trained with more diverse data, e.g., MTG-Humans and MTG-Cards, representations fell short of their counterparts.

\section{Limitations}

As discussed throughout this paper one of the main challenges for unsupervised generative models to face is the diversity within the datasets. Some models can leverage those when conditioned to classes~\cite{odena2017conditional,brock2018large} or text~\cite{xu2018attngan,dash2017tac,souza2020efficient}. Although, in this paper we made it evident that for finetuning purposes it is often paramount to present clearer patterns for the networks to achieve better results. Hence, image and pattern diversity is still a challenge in the field, specially in small data regimes. In addition, we note that even for very small datasets, such as \dataset-Forests (600 instances), it was necessary at least a couple of hours for the model to generate reasonable illustrations -- perhaps the area still lacks in efficient ways to do zero and few-shot transfer. 

\section{Related Work}

To the best of our knowledge, this is the first paper to handle the specific problem of card illustration generation for the Magic card game. Most of the GAN work has been dedicated to generating images in more constrained datasets~\cite{karras2020training,souza2020efficient}. Although, there is some similar related work to some extent. Generating images in a high diversity datasets has been explored by \cite{vavilala2021controlled} in the context of illustrations for Yu-Gi-Oh card game. Training GANs with limited data is a topic that has receiving attention in recent works, such as \cite{karras2020training,tran2020towards,tseng2021regularizing}.

\cite{back2021finetuning}~proposed finetuning StyleGan2 for cartoon face generation with some transfer learning approaches. StylenGan2 is also used in \cite{huang2020unsupervised} for image-to-image translation. AgileGAN \cite{song2021agilegan} uses the pre-trained weights from StyleGan2 to create a framework that can generate high quality stylistic portraits via inversion-consistent transfer learning. Style transfer with StyleGan has also been explored in \cite{abdal2019image2stylegan} and \cite{richardson2021encoding}. 


\section{Conclusion}

In this paper, we proposed \gan~and~\ganada, which are finetuned incarnations of StyleGan2 and StyleGan2 ADA for generating \magic~illustrations. To train those models we introduced~\dataset, a novel dataset that contains 62 thousand illustrations and it is rich in metadata. We show that our models achieve significant results even for very small datasets. Results depicted throughout this paper show that transfer learning is indeed an effective method for generating images even when the source and target domains are rather different from each other -- though models do benefit from pre-trained weights in domains that share structural and aesthetic patterns. We clearly demonstrated that pre-trained features are reused and adapted during tuning. Our experiments also demonstrated that it is paramount to control data diversity for training in such small data regimes. An example of that is that we achieved far better image generative results when training separate models for more specific subsets. It is worth noticing that these models still take a considerable time to generate images even for small datasets. For future work, we aim to improve generative performance of~\gan, specially to generate more complex forms, such as Humans, Animals, Gobblings and the like. To achieve that, we intent to explore Deep Probabilistic Models, as well as allow generation of illustrations conditioned to classes or text.

{\small
\bibliographystyle{IEEEtranS}
\bibliography{main}
}




\clearpage






\end{document}